\documentclass[sigconf]{acmart}
\setcopyright{none}
\copyrightyear{}
\acmDOI{}
\acmISBN{}
\acmConference[KDD Cup Reinforcement Learning]{KDD Cup Reinforcement Learning}{2019}{KDD workshop}

\usepackage[]{algorithm2e}
\AtBeginDocument{%
  \providecommand\BibTeX{{%
    \normalfont B\kern-0.5em{\scshape i\kern-0.25em b}\kern-0.8em\TeX}}}

\begin{document}

\title{Policy Learning for Malaria Control}

\author{Van Bach Nguyen, Belaid Mohamed Karim, Bao Long Vu, J{\"o}rg Schl{\"o}tterer, Michael Granitzer}
\affiliation{%
  \institution{University of Passau}
  \city{Passau}
  \state{Bayern}
  \postcode{94032}
}

\renewcommand{\shortauthors}{LOLs team}

\begin{abstract}
  Sequential decision making is a typical problem in reinforcement learning with plenty of algorithms to solve it. However, only a few of them can work effectively with a very small number of observations. In this report, we introduce the progress to learn the policy for Malaria Control as a Reinforcement Learning problem in the KDD Cup Challenge 2019 and propose diverse solutions to deal with the limited observations problem. We apply the Genetic Algorithm, Bayesian Optimization, Q-learning with sequence breaking to find the optimal policy for five years in a row with only 20 episodes/100 evaluations. We evaluate those algorithms and compare their performance with Random Search as a baseline. Among these algorithms, Q-Learning with sequence breaking has been submitted to the challenge and got ranked 7th in KDD Cup.
\end{abstract}




\maketitle

\section{Introduction}
Malaria is caused by parasites that are transmitted to people through the bites of infected mosquitoes. It is one of the most dangerous diseases in the world. According to WHO, in 2017, nearly half of the world's population was at risk of malaria, among those, there were 219 million cases of Malaria, about 435 000 malaria deaths. Sub Saharan Africa is the home to 92\% of cases and 93\% of deaths\cite{who}. Furthermore, about 450\$ M  is spent in research and development each year to deal with this disease\cite{Moran}. That is why it is also the topic for the KDD challenge 2019 to find out the effective combinations of interventions to prevent malaria infection.

In this challenge, two interventions are considered to control the malaria disease: distributing long-lasting insecticide-treated nets (ITN), and performing indoor residual spraying programs (IRS). Our goal is to determine the most effective policies for five years based on the combinations of these two interventions. the cost-effectiveness of a policy depends on how much we use each intervention among five years. For example, According to a report from IBM \cite{IBM} in some transmission locations, ITNs are the best interventions meanwhile, in Western Kenya, perform IRS in a small proportion of households is more effective, instead of deploying ITNs.

The most important thing that we need to deal with Malaria Problems is the environment. Fortunately, a simulation called OpenMalaria \cite{openmalaria} is created that transforms the real world to the machine world, based on that, we can interact with the environment. OpenMalaria provides a simulation environment that can return a reward each time an agent takes an action. Basing on it, interfaces were created for the challenge, each interface corresponds to an environment. In the KDD cup, two interfaces were provided as training environments and a secret interface was used as the test environment to evaluate solutions.

In this report, we first introduce the challenge as a Reinforcement Learning problem with a limited number of observations. Second, we propose our solutions that include: Random Search as a baseline, Generic Algorithm, Bayesian Optimization and Q-learning with sequence breaking, which is also the final submission, to solve the challenge. Finally, we compare the results of these algorithms and frame future approaches.

\section{Malaria control policy as Reinforcement learning problem}
Malaria challenge is considered as a sequential decision-making problem, which is a typical type of problem in Reinforcement Learning (RL). Therefore, in this section, we will explain some notations in RL before introducing the Malaria challenge.
\subsection{Reinforcement learning}
Reinforcement learning (RL) paradigm enables the autonomous ability for artificial intelligence (AI) machines, in which, we do not need to teach them by providing data or any knowledge for these machines, instead, they learn how to behave by themselves. This idea is inspired by the human development process as we learn new skills not only from teachers but also from a ton of mistakes  (trial-and-error learning). RL has a long history of development. It achieved impressive results in robotics \cite{RLA1,RLA2}, controls \cite{RLA3,RLA4}, and games \cite{RLA5,RLA6}. 
\begin{figure}[h]
  \centering
  \includegraphics[width=\linewidth]{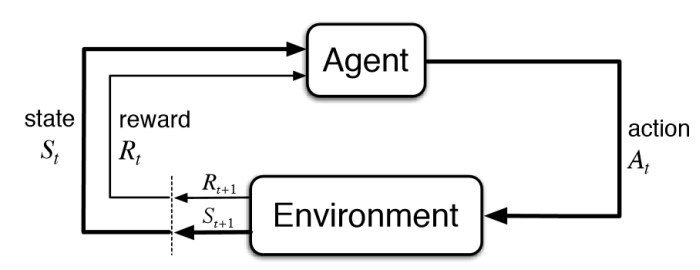}
  \caption{The agent-environment interaction  \cite{RL_Sut})}
  \Description{Hello every one}
  \label{RL_Ar}
\end{figure}
Beside supervised learning and unsupervised learning, reinforcement learning (RL) is an essential paradigm of machine learning. The idea of RL is depicted in figure \ref{RL_Ar}. Unlike supervised learning where a model takes a bunch of examples with ground truth and learns from them, in RL, an \textit{agent} (similar to model) learns by experience through interacting with environment. \textit{Agents} get experience by taking \textit{actions} and receiving \textit{reward} from the \textit{environment}. The goal of agents in RL is maximizing rewards by choosing proper actions in a situation. These situations are called \textit{states}, a strategy to decide an action $A$ in a specific \textit{state} $S$ is called a \textit{policy} $\pi$, i.e. $\pi$ is a function that maps a \textit{state} to an \textit{action}. Agents start from the \textit{initial state}, follow the policy and will stop if they reach the \textit{terminal state}, all \textit{states} that agents go through between the \textit{initial state} and the \textit{terminal state} form  an \textit{Episode}.
\subsection{Malaria Control Challenge}
Malaria Control Challenge is a typical Reinforcement Learning problem that includes:
\begin{itemize}
  \item An Agent: It is the model that we need to build.
  \item An Environment: It is a simulation provided by OpenMalaria. We have two training environments and one final test environment.
  \item Actions: they combine two interventions. Insecticide-Treated Nets (ITNs) and Indoor Residual Spraying (IRS). The domain of the first component is the deployment of nets, which defines the coverage of the population ($a_{ITN} \in [0,1]$). The domain for the second component is the application of seasonal spraying, which defines the proportion of population coverage for this intervention ($a_{IRS} \in [0,1] $). Action space $A$ is continuous space that is constructed through $a_i \in A = \{a_{ITN}, a_{IRS}\}$. A policy maps five years to a set of actions $\{a_1,a_2,a_3,a_4,a_5\}$.
  \item States: they are represented by the year. This means we have five years corresponding to five states. In RL problem, the next state usually depends on the action that agent takes in the current state, however, in this problem, we do not have this kind of state transition, instead, whatever the action is, the state always increases to the next state (next year). An episode always consists of five states.
  \item Rewards: They are modeled as float numbers, the environment will return a reward after each year and a reward for the whole episode. This latter is the sum of the five intermediary rewards (or five states)
\end{itemize}
Our goal is to build an agent that can explore the environment after 20 episodes or 100 evaluations to get the optimal policy that has the highest reward. Reinforcement Learning tasks usually use hundreds or thousands of episodes to train the agent but in this challenge, we only have 20 and our action space is continuous. It is the most difficult part of the challenge. Next section, we will propose some methods to deal with this difficulty.
\section{Solutions}
\subsection{Random Search (Baseline)}
For the Random Search algorithm, the agent just picks 20 arbitrary sets of actions corresponding to 20 episodes, stores the rewards for each episode and finally, chooses the set of actions that has the highest reward. The final policy is: $\pi^*=\arg\max_\pi \pi(s,a)$. This approach is very simple and straightforward. It is only a baseline to find better solutions. However, in our case of very limited episodes, this baseline is already very hard to overcome.
\subsection{Genetic Algorithm}
Genetic Algorithm (GA) is a biologically inspired and population-based algorithm \cite{Holland1992}. It starts from an initial population of policies, called 1st generation and aims to improve through the creation of new generations by simulating natural selection operations.

GA uses the term \textbf{fitness}, which indicates how good a policy is. Fitness can be seen as accuracy in general machine learning algorithms and is normalized to [0, 1] for this problem. Operations are the actions that modify policies in a genetically way. In this paper, 2 operations called \textbf{Mutation} and \textbf{Crossover} are used. The Crossover operation mixes 2 policies randomly or orderly. Here, a policy contains 5 tuples of 2 values, hence when we combine 2 policies using Crossover, some of their tuples will be exchanged and new policies are created. While the Mutation operation changes the value of a tuple randomly, with a given policy, one or more of its tuples' value can be increased or decreased by adding a noise value.

From the current population, which is a set of policies and their rewards, we select candidates for crossover and mutation by Roulette Wheel Selection \textbf{Roulette Wheel Selection} \cite{Goldberg1991} algorithm. Where the probability of selection of a policy \begin{math}b^j\end{math} is determined by dividing its fitness \begin{math}f^j\end{math} to the sum of the fitness of all policies, which is defined in equation (\ref{eq:RWS}).
\begin{equation}
    p^j = \frac{f^j}{\sum_{i=1}^{n} f^i}
    \label{eq:RWS}
\end{equation}

The complete work is as follows: 1st generation will be created stochastically. Then they are pushed into the environment when their rewards come out. Roulette Wheel Selection will choose the 2 best policies and start mating them together using Crossover and Mutation operations. The new child will be pushed to the existing population and the process continue.

\subsection{Bayesian Optimisation (BO)}
Before introducing Bayesian Optimization, we need to present, first, Active Learning.
\subsubsection{What is Active Learning (AL):\newline }
   For certain AI challenges, we can have at our disposal an Oracle/Expert that can answer the targeted question. As an example, we want to create an Optical character recognition (OCR) for Street View House Number. For a supervised learning approach, a human being (Expert) has to label a big number of images with the house number. The trivial way would be to randomly choose images to label. In this case, some images would look very similar and will not help the algorithm learn more. On the other hand, an AL algorithm would estimate which image would improve the current model and then ask the Oracle for help.
\subsubsection{Active Learning in KDD Cup:\newline }
   In the KDD Cup Challenge 2019, the online server represents an Oracle. The agent can ask only 100 queries to the Oracle. The goal of the AL module would be to maximize the potential gained knowledge by the $i$th query based on the result of the previous $i-1$ queries. The AL Algorithm can calculate the distance between all queried policies and a non tested one. If the distance is maximized, we will try to explore the action space. Otherwise, querying around best values will tend to optimize the best policy.
\subsubsection{Bayesian Optimisation and Active Learning:\newline }
Bayesian Optimization is a global optimization algorithm based on Gaussian processes\cite{BOrepo}. It is able to approximate a function using Upper Bound Confidence. the interpolation becomes more and more precise while querying more points using the oracle. The goal is to find the maximum of the function and keep querying around this maxima thanks to its Active Learning module.

\begin{figure}[h]
  \centering
  \includegraphics[width=\linewidth]{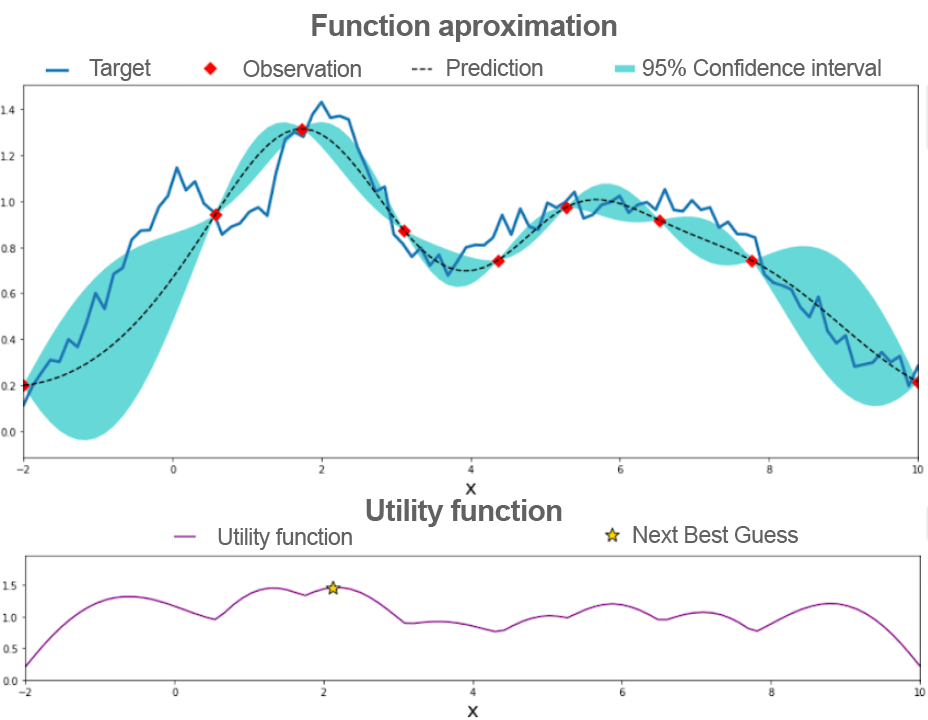}
  \caption{Function approximation using Bayesian Optimization}
  \label{fig:BO01}
\end{figure}

As seen in Figure \ref{fig:BO01}, the algorithm is also able to approximate functions biased with white noise. The Utility function estimates the next best point to query. A parameter Kappa represents the balance between exploration and exploitation (e.g. finding another maximum or querying around the one already found)

To summarize, the advantage of Bayesian Optimisation is that it includes:
\begin{itemize}
    \item Interpolation (even for noisy functions)
    \item Inverse Reinforcement Learning\cite{IRL}
    \item Active Learning
\end{itemize}
More information about global optimization with Gaussian processes can be found here\cite{BOrepo}. 

\subsubsection{Implementation:\newline}

A trivial implementation of BO would be to approximate 5 functions $f_i$ with $i$ representing the year $1 \leq i \leq 5$. The input is the pair of actions on year $i$, and the approximated output is the obtained reward.

\begin{figure}[h]
  \centering
  \includegraphics[width=\linewidth]{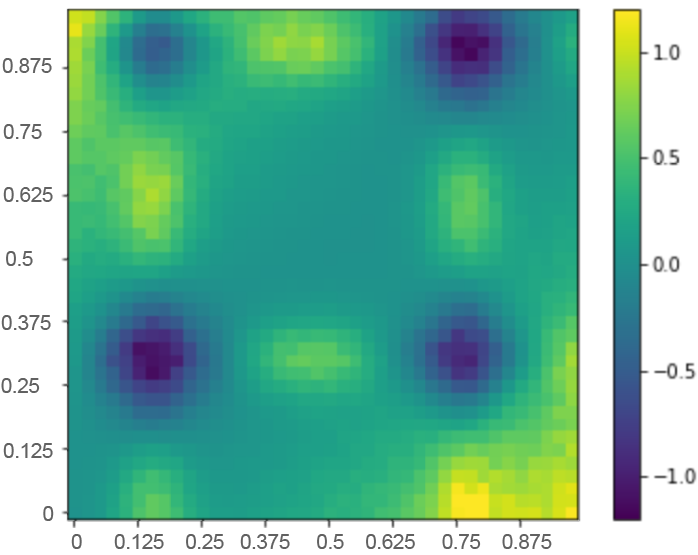}
  \caption{Reward distribution on year 1. The axis represent the pair of actions. The reward value is scaled down by a factor of 1/100} 
  \label{fig:y1}
\end{figure} 
In Figure \ref{fig:y1}, f1 is plotted using the exhaustive search: 1600 (40 by 40) queries were performed to obtain this precise representation. The queried environment was the first one provided for this challenge. We will call it Environment 1 or Sequential Decision Making.

\begin{figure}[h]
  \centering
  \includegraphics[width=0.9\linewidth]{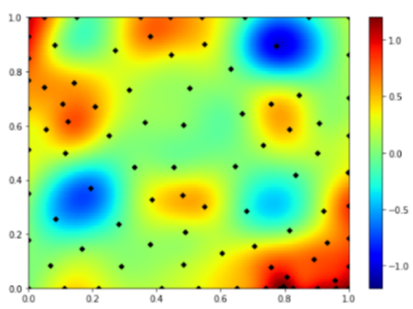}
  \caption{Year 1's reward, approximated using Bayesian Optimization}
  \label{fig:BO02}
\end{figure} 

Figure \ref{fig:BO02} shows the approximation of $f_1$ using BO \cite{BOrepo}. We notice that the shape of the distributions is not well drawn: the Gaussian with low values centered on the point (0.2 , 0.9) is approximated differently than the one centered on the point (0.8 , 0.9). Also, BO was not able to detect the separation between the 2 Gaussian curves with high reward (Figure \ref{fig:BO02}, up left). Despite all these errors, BO is able to detect the maximum of the function and only 89 points were used. Precisely for Environment 1, 25 to 30 queries are enough to find the maximum.

\begin{figure}[h!]
  \centering
  \includegraphics[width=\linewidth]{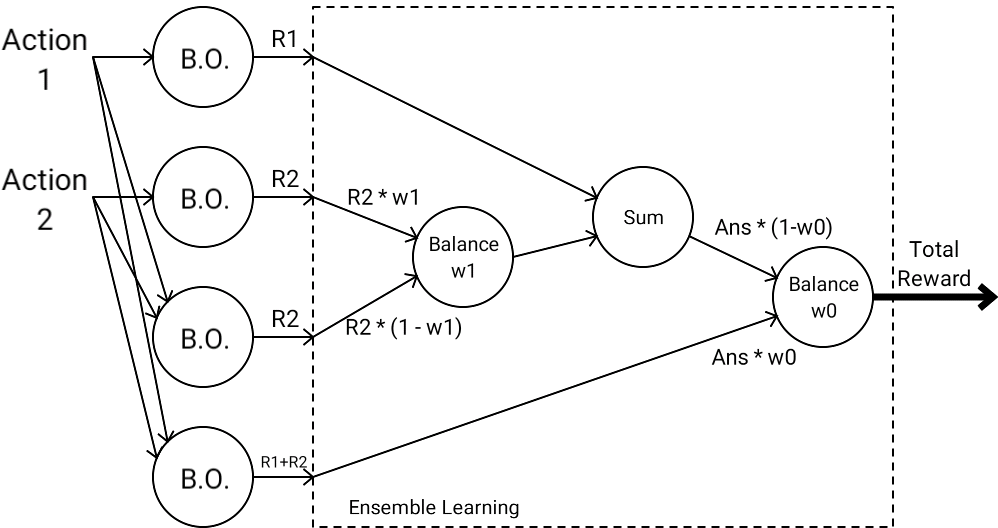}
  \caption{Partial Architecture of the Algorithm BO.3}
  \label{fig:BO_3}
\end{figure}

Since the reward of year $i$ is affected by previous actions, estimating 5 separate functions is not a general solution. To solve this issue, we propose 3 different approaches:

\begin{itemize}
\item Algorithm BO.1 (Figure \ref{fig:BO_1}) uses Bayesian Optimisation for each year: the 5 functions are approximated in a greedy way since that more query points are reserved for the first years.
For example, we first approximate year 1 with 15 points, then we start approximating year 2 based on the optimal action of year 1. Starting from this step, for each new episode, we query year 1 to optimize the current best action, then we query year 2 to explore the action space. Iteratively, we keep going through the years till we reach year 5.
\begin{figure}[h]
  \centering
  \includegraphics[width=0.7\linewidth]{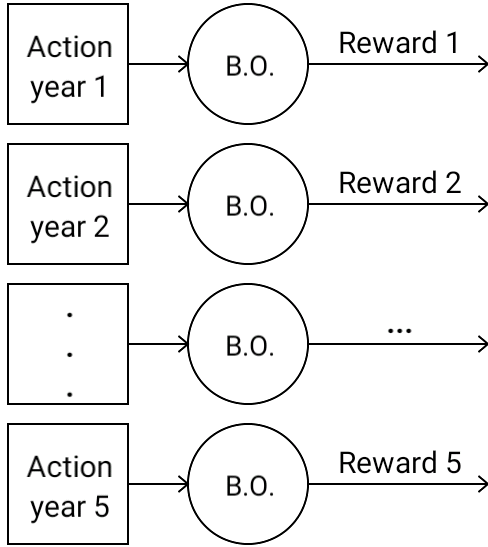}
  \caption{Architecture of the Algorithm BO.1}
  \label{fig:BO_1}
\end{figure}
\item Algorithm BO.2\cite{Mohamed19} uses only one instance of Bayesian Optimisation: the approximated function has ten dimensions as input (the full policy) and one dimension as output (the approximated total reward).
\item Algorithm BO.3 called, "BO with a forward Boosting Network"\cite{Mohamed19}, combine both preceding algorithms thanks to an Ensemble Learning technique: multiple instances of BO with different input and output are first trained. In the second step, a small neural network learns which instance of BO approximates better the environment. based on these weights the agent query the online environment. Upon the reception of the new rewards, the BO instances and the Neural Net. are retrained again.
To keep the Architecture simple, figure \ref{fig:BO_3} represent only the first 2 years. 
\end{itemize}


\subsection{Q-learning with sequence breaking}

In another perspective, we consider the Malaria problem as a sequence of decision-making problems when the reward for a given year depends on all actions that we choose in previous years. Q-Learning was born to solve this kind of problem. In this section, we will introduce the plain Q-learning and the way we modify it in order to apply it to the Malaria problem. This is the solution that we submitted to the KDD competition.
\subsubsection{Plain Q-Learning}

Q-learning \cite{QL}, which is an early breakthrough in reinforcement learning(RL) \cite{RL_Sut}, is defined by
\begin{equation}
    Q(S_t,A_t) \gets Q(S_t,A_t) + \alpha [R_{t+1} + \gamma \max_a Q(S_{t+1},a) - Q(S_t,A_t)]
\end{equation}
This equation expresses that the value of an action in a state $Q(S_t,A_t)$, which evaluates the goodness of an action in a state, will be updated based on: (1) The current value $Q(S_t,A_t)$ that agents have already known, (2) the reward $R_{t+1}$ that agent received after taking action $A_t$, (3) maximum action-value on next state $\max_a Q(S_{t+1},a)$ with discount factor $\gamma$ and (4) learning rate $\alpha$ that defines weights for old and new value that agents just got. If $\alpha = 0$, agents learn nothing, Q-values are unchanged, while if $\alpha = 1$ agents will forget everything from their experience.
$\gamma$ is a discount factor that expresses weight for each reward in a specific time step($t$, ${t+1}$,..), the further time step, the less weight.
All action-values $Q(S_t,A_t)$ are stored in a table called Q-table. This table stores all states with every possible action that agents can take and agents base on this table to choose the best action. The Q-values on this table does not depend on the way agents choose action to take, i.e. Q-learning is off policy.
But in this challenge, with only 100 evaluations (20 episodes) and continuous action space, it is challenging for Q-learning to perform well. 

\begin{figure}[h!]
  \centering
  \includegraphics[width=0.5\linewidth]{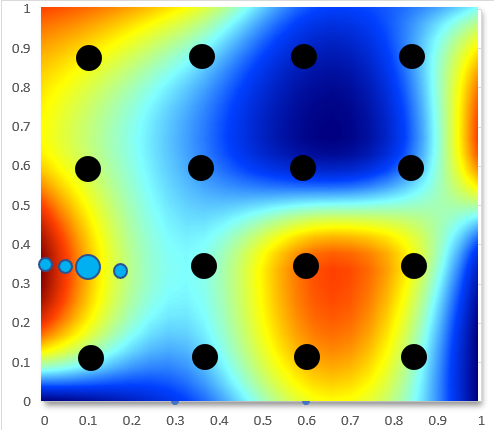}
  \caption{Grid Search: The big blue point is the maximum point after grid search, from that, we query around it to find the best point}
  \Description{Hello every one}
  \label{Grid}
\end{figure}
\subsubsection{Q-learning with sequence breaking}
\begin{figure}[h!]
  \centering
  \includegraphics[width=\linewidth]{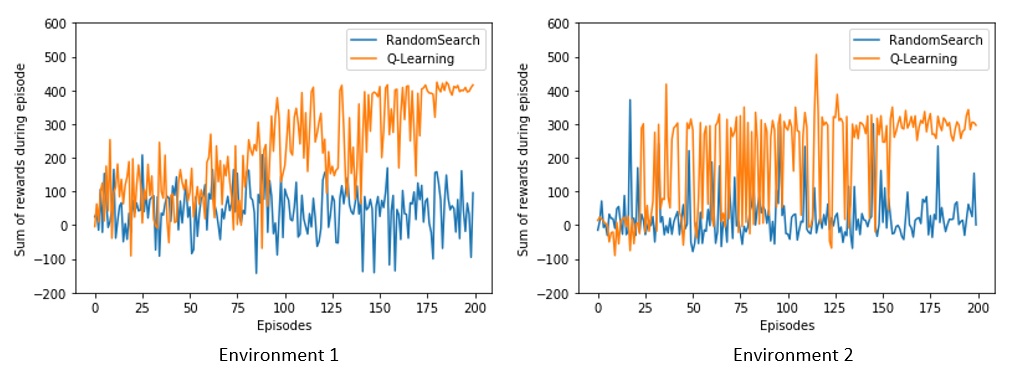}
  \caption{Q-Learning  with 200 episodes result}
  \Description{Hello every one}
  \label{Q-Le}
\end{figure}
Original Q-learning requires discrete action spaces. To satisfy this condition, we only consider actions that have one number in decimal part while the integer part is zero, for example: [0.1,0.2] is a pair of action for a year. There are total 100 pairs of action in our action space. By doing it, we can easily apply algorithms that can only work on discrete action spaces like Q-learning. Furthermore, in these test environments, rewards follow multivariate gaussian distributions, we can quickly find areas that give us high reward. However, with this approach, we cannot reach the global optimal because the best action might be not in the 100 pairs of action in our space. But with limited episodes, it is a good choice to come up with a pair of actions that gives us high rewards.

\begin{algorithm}[h!]
\caption{Q-learning with sequence breaking algorithm}
\SetAlgoLined
\KwResult{Optimal policy $\pi^*$}
 define action space A with resolution = 0.3\;
 take each action in A, store reward in R table\;
 $a_{max} = \arg\max_a(R(a))$\;
 $a_{next}$ = random action around $a_{max}$ with distance = 0.1\;
 \Repeat{4 times }{
  \eIf{$R(a_{next}) > R(a_{max})$}{
        $d = a_{next} - a_{max}$\;
        $a_{max} = a_{next}$\;
        $a_{next} = a_{next} + d$\;
   }{
   $a_{next}$ = random action around $a_{max}$\;
  }

 }
  define action space A with resolution = 0.1\;
  Initialize $\epsilon = 0.8$, $N(s,a) = 1$, $\pi = [], \pi^* =[]$\;
  Initialize $Q(s, a) = 0$, for all $s \in 5 states$, $a \in A$\;
  \For{$e$ in 16 episodes}{
    $\epsilon = 0.8-(e/(16*1.2))$\;
    \eIf{$s_{current} == 1$}
    {
        take $a_{max}$, observe $r, s'$ \;
        $s = s'$\;
    }
    {
    \For{4 years}{
        choose $a$ from A follow $\epsilon-greedy$\;
        take $a$, observe $r, s'$ ;   $\pi$.add($a$)\;
      
        update: $Q(s,a)=Q(s,a) + 1/N(s,a)[R +\gamma \max_a Q(S', a) . Q(S, a)]$\;
        s = s'; $N(s,a) += 1$\;
       
    }
    \If{Reward($\pi$) > Reward($\pi_{max}$)}{
        $\pi^* = \pi$\;
    }
    }
  }
  return $\pi^*$\;
\end{algorithm}
After discretizing our actions space, if we only apply the simple Q-learning, it worked well with a high number of episodes figure \ref{Q-Le}, however, with only 20 episodes, it is not better than the baseline. Therefore, we decided to take some advantages at the beginning of each episode to boost the Q-learning result.  The key idea is: We break the sequence of action into 2 parts: first-year and other years. For the first years, we spent 20 evaluations to find the best pair of action that returns maximum immediate reward this year, ignoring the relationship between the first year and other years. To determine this best action, firstly, we use a grid search with size 4x4 to explore the environment and find areas that have potential high reward. We checked 16 pairs of actions that are a combination of 2 actions from the following list [0.0,0.3,0.6,0.9], for example, [0.3,0.6] or [0.9,0.0], then we choose the pair of actions that give us highest reward among these 16 pairs. Secondly, we use 4 remaining evaluations to exploit the area around the chosen pair of actions. To exploit it, we query random 1 pair action around the current best action, if the chosen action gives higher reward than the best action, we follow that direction, check the pair of action that lies on further part on this direction. For example, if [0.3,0.6] is the best action after grid search, one random action around it will be checked, for example [0.2,0.6]. if this action is better than [0.3 0.6], we will check the next action is [0.1, 0.6], otherwise, we just choose the random action around the best current action, figure \ref{Grid} illustrates the idea.
After coming up with the best action for the 1st year, we will fix it unchanged and apply Q-learning for other years. For Q-learning, we use $\epsilon-greedy$ policy to choose the action with $\epsilon = 0.8$  and learning rate $\lambda = 1/n$ with $n$ is number of action taken times .
The source code for this solution is available on \url{https://github.com/bach1292/KDD_Cup_2019_LOLS_Team.git}

Another simple approach is that we can apply a grid search for all five years, ignore the relation between each year. This completely breaks the sequence relation of five years and only applies a grid search. The result was even better than combining with Q-learning, however, due to the restriction of the competition, we could not submit it.

\section{Results}
\subsection{Genetic Algorithm}
\begin{figure}[H]
  \centering
  \includegraphics[width=0.75\linewidth]{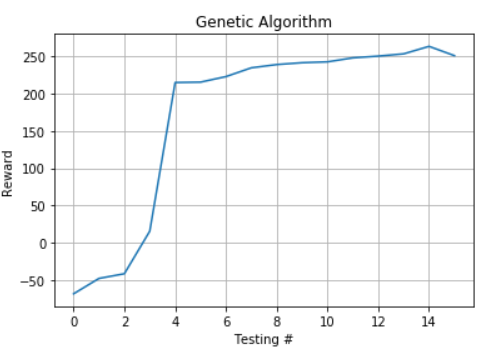}
  \caption{Genetic Algorithm with noise value of 0.05}
  \Description{}
  \label{fig:GA_Result}
\end{figure}
Mutation operation is the main way to control the exploration-exploitation tradeoff, the lower the noise's value, the learning are more exploitation-based.

With a limitation of only 100 testing times, here we use a noise value of 0.05, which is nearly pure exploitation, with such configuration Genetic Algorithm can learn new policies that can yield around 250.00 worth of rewards on the \begin{math}2^{nd}\end{math} environment. See more on figure \ref{fig:GA_Result} and table \ref{tab:a}.


\subsection{Bayesian Optimization}

\begin{table}[h]
  \caption{Comparing the results of BO.1 and BO.2}
\begin{center}
\begin{tabular}{|c|cccccccc|}
      \hline
      \multicolumn{1}{c}{} & & & & & & & &  \\[\dimexpr-\normalbaselineskip-\arrayrulewidth]
      \textbf{Algorithms} &\multicolumn{8}{c|}{Average rewards}\\
      \hline
      \textbf{} & \multicolumn{4}{c|}{Env. 1} & \multicolumn{4}{c|}{Env. 2}\\
      \hline
      \textbf{Random Search (5 years)} & \multicolumn{2}{c|}{161.20} & \multicolumn{2}{c|}{100\%} & \multicolumn{2}{c|}{158.991} & \multicolumn{2}{c|}{100\%} \\
    
      \hline
      \hline
      \textbf{BO.1: 5 indep. B.O.} & \multicolumn{2}{c|}{250} & \multicolumn{2}{c|}{155\%} & \multicolumn{2}{c|}{500} & \multicolumn{2}{c|}{315\%} \\
      \hline
      \textbf{BO.2: 10-dim B.O. } & \multicolumn{2}{c|}{400} & \multicolumn{2}{c|}{248\%} & \multicolumn{2}{c|}{200} & \multicolumn{2}{c|}{126\%} \\
      \hline
    \end{tabular}
    \end{center}
  \label{tab:BO5yearsX}
    \end{table}

Concerning the first implementation (BO.1), only a few queries are left for episodes reaching the terminal state. Therefore, year 5's action space is not well explored comparing to the first years but we generally get a high total reward.
For Environment 1, the maximum total reward is around 110 for each year. It is approximated using the simple exhaustive search algorithm used before.
As shown in table \ref{tab:BO5yearsX}, BO.1 obtains a score close to the maximum in Environment 2. On the other hand, BO.2 obtains a score close to the maximum in Environment 1.
This is explained by the nature of each environment:
\begin{itemize}
\item In Environment 1, the reward of year $i$ depends only on 2 actions: the current one and the one before. By breaking the sequence for 5 years, we learn the best action on year $i$ before proceeding with year $i+1$. Hence, the algorithm BO.1 can optimize year $i+1$'s reward for a fixed action of year $i$. 
\item In Environment 2, the reward of year $i$ depends on current action and all previous actions. A 10-dimensional function - as presented in BO.2 - is able to catch this relation and output the best policy.
\end{itemize}

For BO.3, only a proof of concept was implemented: the algorithm can learn the best policy only for the first 2 years. Results are shown in table \ref{tab:BO2years}. 

\begin{table}[h]
  \caption{Results of BO.3.}
\begin{center}
\begin{tabular}{|c|cccccccc|}
      \hline
      \multicolumn{1}{c}{} & & & & & & & &  \\[\dimexpr-\normalbaselineskip-\arrayrulewidth]
      \textbf{Algorithms} &\multicolumn{8}{c|}{Average rewards}\\
      \hline
      \textbf{} & \multicolumn{4}{c|}{Env. 1} & \multicolumn{4}{c|}{Env. 2}\\
      \hline
      \textbf{Random Search (2 years)} & \multicolumn{2}{c|}{35} & \multicolumn{2}{c|}{100\%} & \multicolumn{2}{c|}{105} & \multicolumn{2}{c|}{100\%} \\
    
      \hline
      \textbf{BO.3: Boosting Network} & \multicolumn{2}{c|}{70} & \multicolumn{2}{c|}{200\%} & \multicolumn{2}{c|}{210} & \multicolumn{2}{c|}{140\%} \\
      \hline
    \end{tabular}
    \end{center}
  \label{tab:BO2years}
    \end{table}

\begin{figure}[h]
  \centering
  \includegraphics[width=0.9\linewidth]{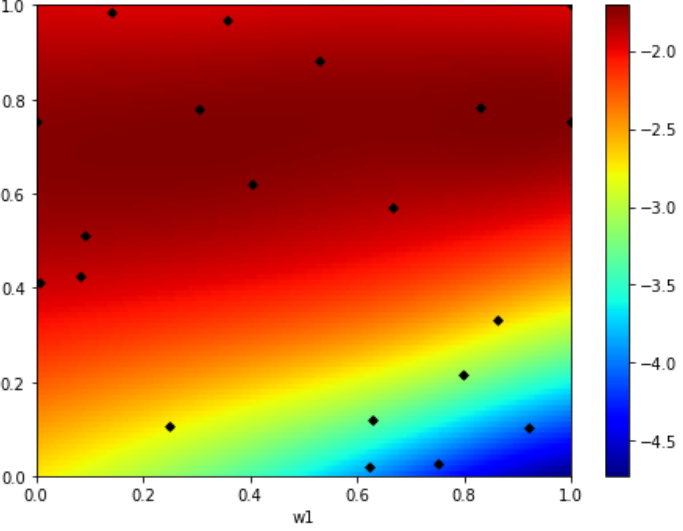}
  \caption{Mean Square error of the predicted reward w.r.t. the 2 weights [w1 , w2] of the boosting network.  }
  \label{fig:BO_3weight}
\end{figure}

Compared to the first two algorithms, BO.3 is able to achieve stable results since it is a balance between the two other solutions. This improvement is achieved by learning the two weights $w_0$ and $w_1$. As example, In the figure \ref{fig:BO_3weight} the best weights are ($w_0$: 0.78, $w_1$: 0.82). 5 to 12 episodes are usually used before reaching the perfect weights. Therefore, BO.3 can not outperform the two other algorithms. 

\subsection{Q-learning with sequence breaking algorithm}

To evaluate the result, the algorithm needs to be run for 10 times, 20 episodes each time and no transferring knowledge between each time, and then we take the average of these 10 runs. This evaluation makes sure that the algorithms produce better result than the baseline. The figure \ref{en1_rs} below shows the comparisons between each algorithm.
As we can see, simple Q-learning cannot perform much better than random search, in some runs, it is even worse than the baseline. Meanwhile, both Q-learning with first-year breaking or completely breaking sequences approach outperformed the others. The best approach is completely breaking sequences, however, we were not allowed to submit it to the competition, this is just our proposed solution. The table \ref{tab:a} shows the average rewards for each algorithm and comparison between them.
\begin{figure}[h!]
  \centering
  \includegraphics[width=\linewidth]{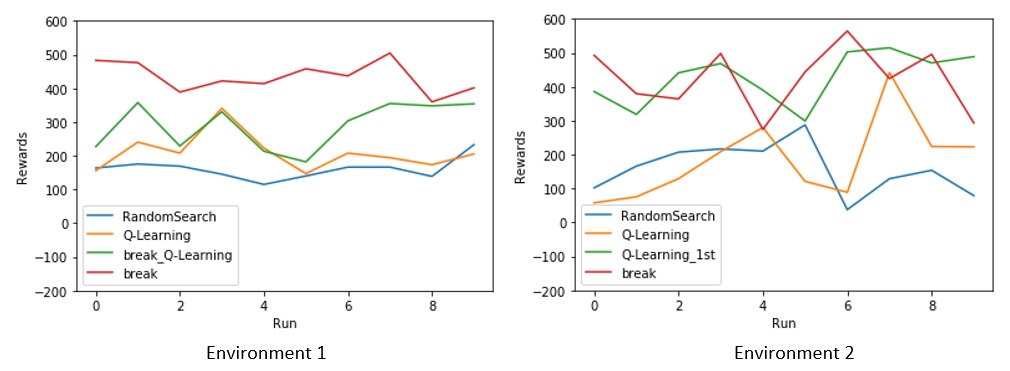}
  \caption{Result comparison}
  \Description{Hello every one}
  \label{en1_rs}
\end{figure}
\begin{table}[ht]

\caption{Comparison between algorithms}
\begin{center}
\begin{tabular}{|c|cccccccc|}
      \hline
      \multicolumn{1}{c}{} & & & & & & & &  \\[\dimexpr-\normalbaselineskip-\arrayrulewidth]
      \textbf{Algorithms} &\multicolumn{8}{c|}{Average rewards}\\
      \hline
      \textbf{} & \multicolumn{4}{c|}{Environment 1} & \multicolumn{4}{c|}{Env. 2}\\
      \hline
      \textbf{Random Search (Baseline)} & \multicolumn{2}{c|}{161.20} & \multicolumn{2}{c|}{100\%} & \multicolumn{2}{c|}{158.991} & \multicolumn{2}{c|}{100\%} \\
      \hline
      \textbf{Break. Seq. for 5 years} & \multicolumn{2}{c|}{434.36} & \multicolumn{2}{c|}{290\%} & \multicolumn{2}{c|}{423.019} & \multicolumn{2}{c|}{267\%} \\
      \hline
      \textbf{Genetic Algorithm} & 
      \multicolumn{2}{c|}{148.18} & \multicolumn{2}{c|}{91\%} & \multicolumn{2}{c|}{262.83} & \multicolumn{2}{c|}{163\%} \\
      \hline
      \textbf{Q-learning} & 
      \multicolumn{2}{c|}{209.51} & \multicolumn{2}{c|}{130\%} & \multicolumn{2}{c|}{185.055} & \multicolumn{2}{c|}{117\%} \\
      \hline
      \textbf{1st year break+Q-Learning} & \multicolumn{2}{c|}{289.93} & \multicolumn{2}{c|}{180\%} & \multicolumn{2}{c|}{428.01} & \multicolumn{2}{c|}{269\%} \\
      \hline
    \end{tabular}
    \end{center}
    \label{tab:a}
    \end{table}

Both sequences breaking algorithms, which use sequence breaking for the first year or whole five years, have common advantages and disadvantages. We can see they work better than baseline and traditional Reinforcement Learning algorithms like plain Q-learning with a short run, they are also quite simple to implement. For long runs with enough episodes, simple Q-learning may be the better choice. The disadvantage of these approaches is that if the maximum reward of the first year leads to a very bad reward in the next four years, it will be a disaster. Furthermore, because we discretized the action space, we cannot obtain the global optimum. However, with limited observations, these algorithms are not bad choices.
\section{Conclusion}
Sequence decision making is always an interesting task for Reinforcement Learning that can be applied to many real-world problems such as Malaria control. However, with a very small number of observations, this task was very hard to solve, traditional Reinforcement Learning algorithms cannot overcome the baseline. This paper has introduced enhanced algorithms that can deal with this difficulty and increase the performance of traditional methods. We hope that our solutions would help in controlling the spread of Malaria disease or at least in bringing up helpful ideas.

\bibliographystyle{ACM-Reference-Format}
\bibliography{KDD}


\end{document}